\documentclass{article}
\usepackage{spconf,amsmath,graphicx}

\usepackage{enumitem}
\setlist{nosep, leftmargin=14pt}

\usepackage{amsmath}
\usepackage{amssymb}
\usepackage{hyperref}

\title{Unsupervised Representation Learning from Pathology Images with Multi-directional Contrastive Predictive Coding}

\name{Jacob Carse$^{\star}$ \qquad Frank Carey$^{\dagger}$ \qquad Stephen McKenna$^{\star}$ }
\address{$^{\star}$Computer Vision and Image Processing, School of Science and Engineering, \\University of Dundee, Dundee, UK\\$^{\dagger}$Department of Pathology, Ninewells Hospital and Medical School, Dundee, UK}

\begin{document}

\maketitle

\begin{abstract}

    Digital pathology tasks have benefited greatly from modern deep learning algorithms. However, their need for large quantities of annotated data has been identified as a key challenge. This need for data can be countered by using unsupervised learning in situations where data are abundant but access to annotations is limited. Feature representations learned from unannotated data using contrastive predictive coding (CPC) have been shown to enable classifiers to obtain state of the art performance from relatively small amounts of annotated computer vision data. We present a modification to the CPC framework for use with digital pathology patches. This is achieved by introducing an alternative mask for building the latent context and using a multi-directional PixelCNN autoregressor. To demonstrate our proposed method we learn feature representations from the Patch Camelyon histology dataset. We show that our proposed modification can yield improved deep classification of histology patches.
    
\end{abstract}

\begin{keywords}
    Medical Imaging, Digital Pathology, Representation Learning, Semi-Supervised Learning
\end{keywords}

\section{Introduction}
\label{sec:intro}

    \begin{figure}[!t]
        \begin{minipage}[b]{.48\linewidth}
            \centering
            \centerline{\includegraphics[width=4.0cm]{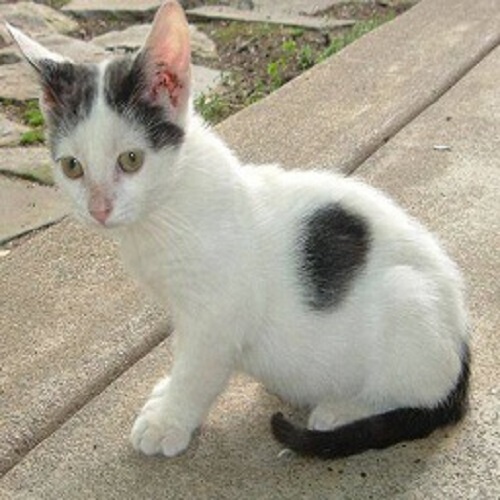}}
            \centerline{(a)}\medskip
        \end{minipage}
        \hfill
        \begin{minipage}[b]{0.48\linewidth}
            \centering
            \centerline{\includegraphics[width=4.0cm]{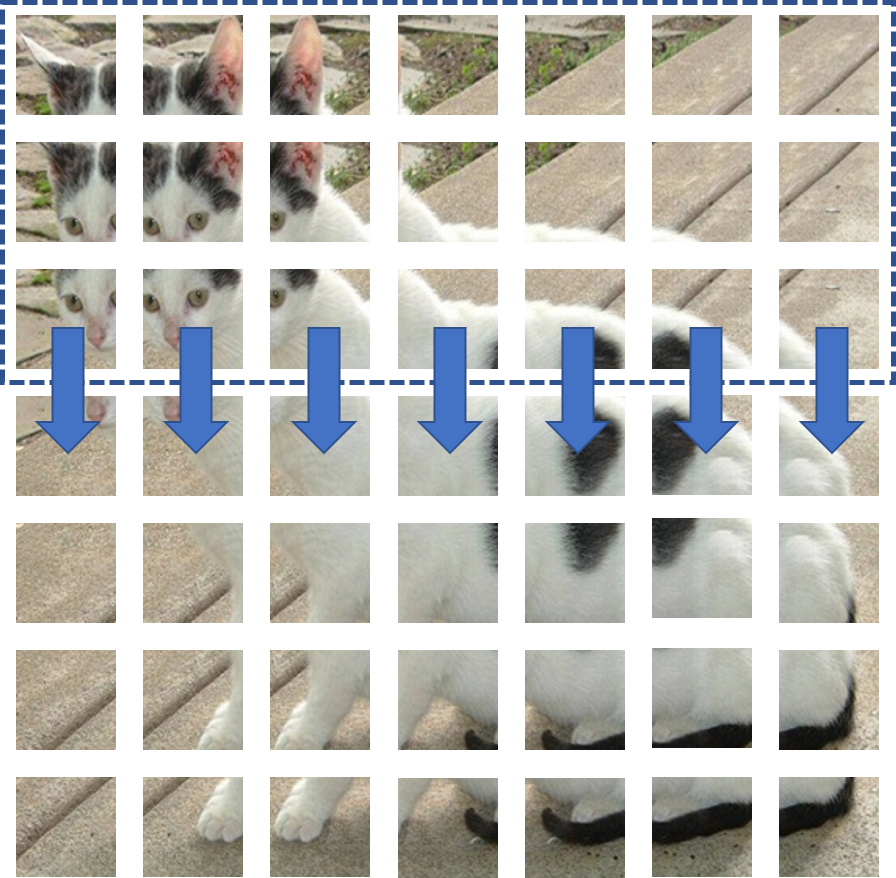}}
            \centerline{(b)}\medskip
        \end{minipage}
        \\
        \begin{minipage}[b]{.48\linewidth}
            \centering
            \centerline{\includegraphics[width=4.0cm]{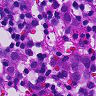}}
            \centerline{(c)}\medskip
        \end{minipage}
        \hfill
        \begin{minipage}[b]{.48\linewidth}
            \centering
            \centerline{\includegraphics[width=4.0cm]{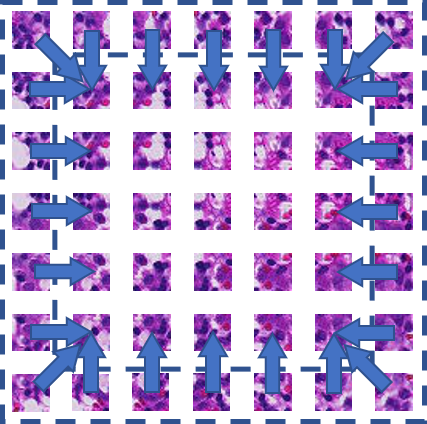}}
            \centerline{(d)}\medskip
        \end{minipage}
        \caption{(a) An example image from the ImageNet dataset~\cite{deng2009imagenet}. (b) Extracted overlapping patches with those used to produce context and autoregressor direction highlighted. (c) An example image from the Patch Camelyon dataset~\cite{veeling2018rotation}. (d) Extracted overlapping patches with those used to produce context and autoregressor direction highlighted.}
        \label{fig:example_cpc_patches}
    \end{figure}

    Modern deep learning algorithms have been shown to improve performance for digital pathology tasks such as nuclei detection and disease classification~\cite{litjens2017survey}. They have been able to achieve such results by jointly learning deep representations and discriminative classifiers or regressors. This end-to-end training allows them to learn feature representations tailored to the given task. However, this approach requires more annotated data than traditional machine learning methods to obtain good generalization. This need for large annotated datasets has been identified as a challenge for digital pathology~\cite{madabhushi2016image}, as well as other medical image analysis domains~\cite{litjens2017survey}. This study focuses on the use of unsupervised representation learning for digital pathology images. The learned representations can potentially be used in multiple tasks to improve generalization, decrease data dimensionality, improve computational performance, and initialize deep supervised learning models when access to annotated data is limited. 
    
    Transfer learning has been performed using the weights of a deep encoder used for generating representations to initialize another model~\cite{weiss2016survey}. Such an approach can reduce the need for large annotated datasets as the model is initialized with weights that can produce domain-representative features, having already been trained on a large pool of unannotated data. 
    
    Contrastive predictive coding (CPC) is a state-of-the-art method for unsupervised representation learning~\cite{oord2018representation}. It learns data representations using an autoregressor to predict future data representations in a sequence. A CPC model is trained using a loss function made from a combination of noise-contrastive estimation and importance sampling where the density ratio between each sample and their representation can be preserved by classifying a positive sample from multiple negative samples. To apply this method to images, van den Oord et al.~\cite{oord2018representation} split each image into overlapping patches and used an encoder to encode each patch to produce a matrix of feature representations. A mask was then applied to this matrix so that an autoregressor could only 'see' the top few rows of it. The autoregressor was then used to predict the representations of the masked patches from this context available to it. An example of this framework being applied to an image can be seen in Fig.~\ref{fig:example_cpc_patches}(b). 
    
    CPC has been used to achieve data-efficient results for object detection and Imagenet classification tasks after modifying model capacity, introducing layer normalization, modifying prediction directions and adding patch-based augmentations~\cite{henaff2019data}. Although autoregressor predictions were made in multiple directions these were done individually. We argue that a different approach is appropriate when the orientation of the images is arbitrary so does not carry useful information. This is the case for many (though not all) digital pathology tasks, in contrast to images in typical computer vision tasks. 
    
    In this paper we propose using CPC with an alternative mask for building latent context and a new extension to the autoregressor PixelCNN for multi-directional prediction (Fig.~\ref{fig:example_cpc_patches}(d)). We demonstrate, using the PatchCamelyon dataset~\cite{veeling2018rotation} (derived from the Camelyon16 dataset~\cite{litjens20181399}), that classification can be performed with less annotated data using representations learned in this way. 
    
\section{Related Work}
\label{sec:related_work}

    In recent years, numerous advances in unsupervised visual representation learning have been proposed. To better learn representations specifically for classification tasks, self-supervised learning methods such as RotNet~\cite{gidaris2018unsupervised} and deep clustering~\cite{caron2018deep} are often used. RotNet uses a convolutional neural network (CNN) to predict geometric transforms of an input image. The deep clustering methods trains a CNN encoder by clustering feature representations and then classifies the images using the clusters as pseudo-labels.
    
    Momentum contrast uses a dynamic dictionary of keys to train an encoder~\cite{he2020momentum}. This encoder is trained using a contrastive loss comparing a query image's representation to the key representations from the dynamic dictionary made up of a queue of data samples. The authors state that because their dictionary is large and dynamic, they can sample high-dimensional visual space effectively and as a result show competitive results. They show that their learned feature representations can be used to pre-train encoders for tasks such as classification, detection and segmentation.
    
    In digital pathology, transfer learning is commonly used for various tasks~\cite{srinidhi2020deep} and helps to speed up convergence of deep learning~\cite{bayramoglu2016transfer}. Some studies have used representation learning to initialise a CNN when the learning task may be too difficult otherwise due to a lack of annotated data~\cite{hou2016automatic}. This could also extend to active learning; when annotations are limited features can be hard to learn~\cite{carse2019active}. 
    
    Instead of using features trained on general computer vision data, some authors have sought to learn more general histology features for cell-level tasks~\cite{hu2018unsupervised}. This has been done by training a unified generative adversarial network with a modified loss function for cell-level representation learning. Another method for learning general histology features used multi-scale convolutional sparse coding, seeking to jointly learn features at different scales with enforced scale-specificity~\cite{chang2017unsupervised}. 
    
\section{Contrastive Predictive Coding in Computer Vision}
\label{sec:cpc}

    A CPC model is trained by predicting future representations from past representations, enabling it to learn `slow' features that effectively represent an input data distribution~\cite{henaff2019data,oord2018representation}. This is done by jointly training an encoder, \(g_{en}\), and an autoregressor, \(g_{ar}\). The encoder encodes each element, \(x_t\), of input sequence \(x\) into a latent representation, \(z_t = g_{en}(x_t)\). The autoregressor then summarizes part of the latent representation sequence \(z_{\le t}\) into a latent context representation \(c_t = g_{ar}(z_{\le t})\). A density ratio \(f\) can be modeled that can preserve the mutual information between \(x_{t+k}\) and \(c_t\) (Equation~(\ref{eq:density_ratio})).
    
    \begin{equation}
        f_k(x_{t+k}, c_t) \propto \frac{p(x_{t+k}|c_t)}{p(x_{t+k})}
        \label{eq:density_ratio}
    \end{equation}
    
    \noindent The InfoNCE loss function used to jointly train the encoder and autoregressor is based on noise-contrastive estimation and importance sampling. There is no way to evaluate \(p(x)\) or \(p(x|c)\) directly so instead a cross entropy loss is used to classify positive samples of \(z\) from random negative samples. Details of the function \(f\) and the InfoNCE loss can be found in~\cite{oord2018representation}.

    
     This CPC method has been applied to computer vision by taking overlapping patches from each image as in Fig.~\ref{fig:example_cpc_patches}. Each of these patches is then encoded and an autoregressor is used to produce a context vector from the patch representations at the top of the image (as illustrated in Fig.~\ref{fig:example_cpc_patches}(b) where the top 3 rows of patches were used~\cite{oord2018representation}). Each column of the image is then treated as a sequence with the context vector from the top of the image used to predict the remaining patch representations below. CPC has been used to achieve data-efficient results on computer vision datasets such as Imagenet~\cite{deng2009imagenet}. 

\section{Multi-Directional Contrastive Predictive Coding}
\label{sec:proposed_modification}

    The treatment of columns of the representation matrix as individual sequences (as in the method described in Section~\ref{sec:cpc}) can negatively impact performance when working with images such as histology patches in which overall image rotation is irrelevant. Although the orientation of certain histology whole slide images can be biologically meaningful, the orientation of image patches such as those used in our experiments (images of lymph nodes with breast cancer metastases) is not. In such cases, the autoregressor can struggle to predict patch representations from the provided context because the vertical image axis is arbitrary, unlike in Imagenet where it correlates with the direction of gravity acting upon the image content. We make two modifications inspired by image in-filling to address this limitation: an alternative latent mask for producing a context vector, and a modified PixelCNN~\cite{oord2016pixel} for multi-directional context building. We propose multi-directional CPC, utilizing these two modifications, to more effectively learn representations from images when image rotation is uninformative.
    
\begin{figure}[!tb]
        \begin{minipage}[b]{1.0\linewidth}
            \centering
            \centerline{\includegraphics[width=8.3cm]{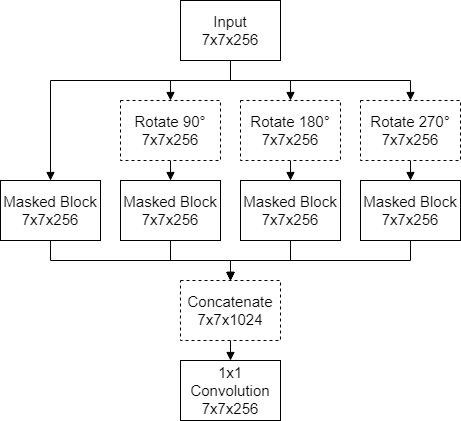}}
        \end{minipage}
        \caption{Architecture of a Multi-Directional Masked Block}
        \label{fig:multi-directional_masked_block}
    \end{figure}

    We first present a modified version of PixelCNN that will be used as the autoregressor. This method replaces each masked convolutional block of the PixelCNN architecture with a multi-directional masked block (Fig.~\ref{fig:multi-directional_masked_block}). Each multi-directional masked block takes a single input image patch and rotates through 90, 180 and 270 degrees, resulting in four versions of the input. A masked block, as described in~\cite{oord2016pixel}, is then applied to each of them. The four outputs from the masked blocks are then concatenated and put through a final 1x1 convolutional layer for the purpose of dimensionality reduction.
    
    We use this multi-directional autoregressor to learn a latent context from multiple directions at once. To take advantage of this we introduce an alternative latent mask inspired by in-filling. With this mask (illustrated in Fig.~\ref{fig:example_cpc_patches}(d)), the autoregressor only has access to patch representations around the perimeter of the images. The autoregressor predicts the patch representations of the central patches. This means that images where rotation is unimportant can be better represented with features learned using CPC.

\section{Experiments and Results}
\label{sec:experiments}

    \begin{figure*}[!t]
        \centering
        \includegraphics[width=0.85\textwidth]{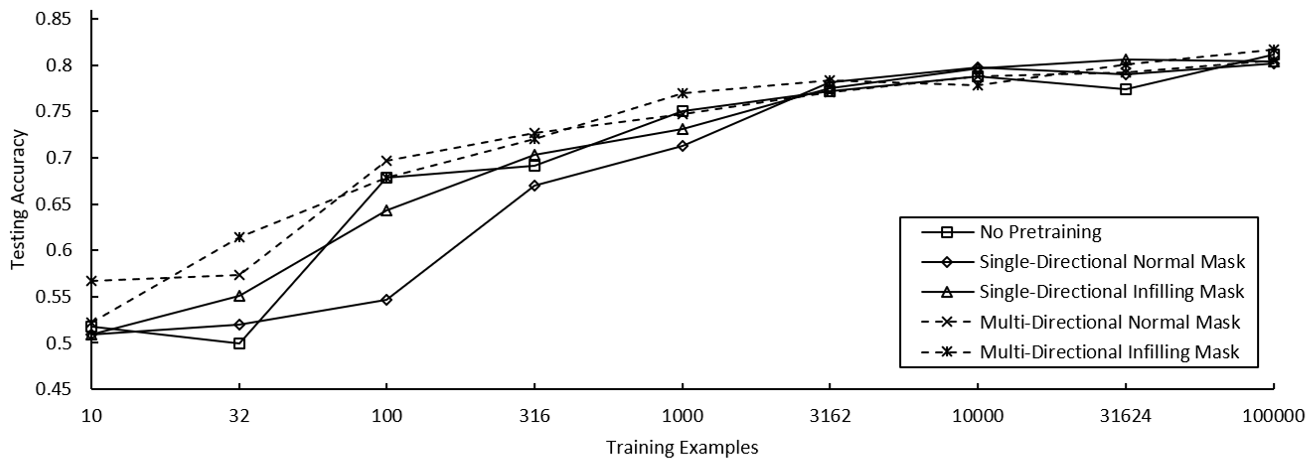}
        \caption{Mean test accuracies of the CNN classifiers.}
        \label{fig:cnn_training}
    \end{figure*}
    
    \begin{table*}[t]
        \resizebox{\textwidth}{!}{%
        \begin{tabular}{l|l|l|l|l|l|l|l|l|l}
            Training Examples & 10 & 32 & 100 & 316 & 1000 & 3162 & 10000 & 31624 & 100000 \\ \hline
            No Pretraining & 0.518 (0.041) & 0.500 (0.000) & 0.679 (0.014) & 0.691 (0.053) & 0.750 (0.014) & 0.772 (0.017) & 0.788 (0.005) & 0.773 (0.003) & 0.811 (0.004) \\
            Single-Directional Normal Mask & 0.509 (0.029) & 0.520 (0.045) & 0.546 (0.046) & 0.670 (0.029) & 0.713 (0.022) & 0.781 (0.014) & 0.798 (0.003) & 0.790 (0.009) & 0.802 (0.008) \\
            Single-Directional Infilling Mask & 0.509 (0.055) & 0.551 (0.042) & 0.643 (0.049) & 0.703 (0.014) & 0.731 (0.012) & 0.775 (0.012) & 0.797 (0.028) & 0.806 (0.014) & 0.804 (0.020) \\
            Multi-Directional Normal Mask & 0.566 (0.071) & 0.573 (0.068) & 0.696 (0.065) & 0.727 (0.018) & 0.747 (0.002) & 0.771 (0.015) & 0.788 (0.018) & 0.792 (0.007) & 0.805 (0.012) \\
            Multi-Directional Infilling Mask & 0.522 (0.045) & 0.614 (0.050) & 0.678 (0.004) & 0.720 (0.035) & 0.769 (0.012) & 0.784 (0.009) & 0.778 (0.013) & 0.801 (0.013) & 0.816 (0.020)
        \end{tabular}%
        }
        \caption{Mean test accuracies of the CNN classifiers (standard deviations in parentheses).}
        \label{tab:cnn_results}
    \end{table*}

    Our experiments used the Patch Camelyon dataset comprised of 327,680 96x96 pixel non-overlapping patches~\cite{veeling2018rotation}. These patches are from scans of lymph node sections in the Camelyon16 dataset~\cite{litjens20181399}. They are annotated with a binary label indicating the presence or absence of metastatic tissue. One reason we selected this dataset was that the small image size allows for larger batch sizes and faster training so that we were able to run experiments using a single Nvidia GeForce RTX 2080 Ti.
    
    We trained four CPC models with different combinations of single- or multi-directional autoregressors, and top-down or in-filling latent masks. To test the effectiveness of the models, we used the trained encoder weights to initialize the weights of CNN classifiers. For each CPC model, nine classification models were trained on subsets of the training data with their labels made available; these labelled subsets ranged in size from 10 to 100,000 images.
    
    \begin{figure}[!b]
        \begin{minipage}[b]{1.0\linewidth}
            \centering
            \centerline{\includegraphics[width=8.3cm]{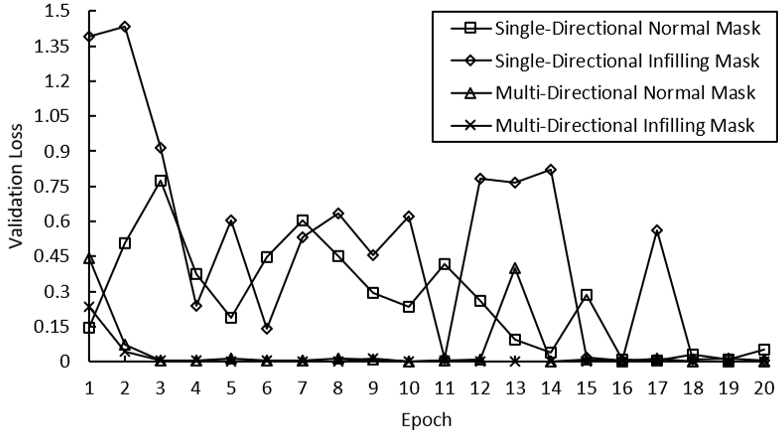}}
        \end{minipage}
        \caption{Validation losses during CPC model training.}
        \label{fig:cpc_training}
    \end{figure}
    
    Our CPC models split each input image into overlapping patches (24x24 patches overlapping by 12 pixels) and used a ResNeXt~\cite{xie2017aggregated} with 101 layers as the encoder followed by an additional convolutional layer to produce a 128-dimensional feature vector for each patch. The autoregressor is comprised of 6 masked convolutional blocks to produce the context vector and predict the masked feature vectors. The loss for the training of the CPC model uses 16 randomly selected images as negative samples in the loss function.

    When training the CPC and CNN models, random rotations (multiples of 90 degrees) as well as horizontal and vertical flips were applied to augment the training data. 20\% of the training data was used as a validation set (58,982 images) for early stopping. We trained the CPC models for 20 epochs using a learning rate of \(1e^{-4}\) and a batch size of 16. The CNN classifiers had a hidden layer of 256 neurons between the encoder and the output layer. They were trained for 50 epochs with the Adam optimizer~\cite{kingma2014adam}, a learning rate of \(1e^{-4}\), and a batch size of 64. Our experiments can be found on our GitHub Repository\footnote{\url{github.com/UoD-CVIP/Multi_Directional_CPC_Histology}}.
    
    Each CPC model took an average of 90 hours to train. Fig.~\ref{fig:cpc_training} plots validation loss at each epoch and clearly suggests that use of a multi-directional autoregressor was more effective at reducing the InfoNCE loss than a top-down autoregressor. The in-filling style in combination with the multi-directional autoregressor stabilized CPC training.
    
    Fig.~\ref{fig:cnn_training} shows average accuracies on a held out testing set (32,768 images) for CNN classifiers trained using the different CPC models' weights for initialization. A baseline using no pre-training is also included. CNNs trained on small amounts of annotated data were able to achieve higher accuracies when a multi-directional autoregressor had been used.  In contrast, standard CPC struggled to learn a representation suitable for initialising a CNN classifier (transfer learning), sometimes lowering accuracy compared to classifiers trained using random weight initialisation (no pre-training). 
    
\section{Conclusion}
\label{sec:conclusions}
    
    We have shown that the original CPC architecture does not perform well on a digital pathology patch classification task. We proposed a multi-directional modification to CPC that achieved better results, and improved classification accuracies when annotated data were limited. Our experiment illustrates that algorithms based on an assumption of image directionality (such as is present in Imagenet) will not necessarily perform well on images without such directionality. We suppose that this will also be true for other visual tasks where image orientation is unimportant, as is the case in multiple biomedical imaging settings.

\section{Compliance with Ethical Standards}
\label{sec:ethics}
    This research study was conducted retrospectively using human subject data made available in open access by~\cite{litjens20181399}. Ethical approval was not required as confirmed by the license attached with the open access data.

\section{Acknowledgments}
\label{sec:acknowledgments}
    This work was supported by the UK Engineering and Physical Sciences Research Council (EPSRC Training Grant EP/N509632/1). The authors are grateful to the Computer Vision and Image Processing research group for helpful motivating discussions. 

\bibliographystyle{IEEEbib}
\bibliography{refs}

\end{document}